\documentclass[letterpaper]{article} % DO NOT CHANGE THIS
\usepackage{aaai2026}  % DO NOT CHANGE THIS
\usepackage{times}  % DO NOT CHANGE THIS
\usepackage{helvet}  % DO NOT CHANGE THIS
\usepackage{courier}  % DO NOT CHANGE THIS
\usepackage[hyphens]{url}  % DO NOT CHANGE THIS
\usepackage{graphicx} % DO NOT CHANGE THIS
\usepackage{natbib}  % DO NOT CHANGE THIS AND DO NOT ADD ANY OPTIONS TO IT
\usepackage{caption} % DO NOT CHANGE THIS AND DO NOT ADD ANY OPTIONS TO IT
\usepackage{algorithm}
\usepackage{algorithmic}
\usepackage{amsmath}

\usepackage{subcaption} % Added to support subfigure environment
\usepackage{newfloat}
\usepackage{listings}
\usepackage{url}            % simple URL typesetting
\usepackage{booktabs}       % professional-quality tables
\usepackage{amsfonts}       % blackboard math symbols
\usepackage{nicefrac}       % compact symbols for 1/2, etc.
\usepackage{microtype}      % microtypography

\usepackage{todonotes}
\usepackage{caption}
\usepackage{graphicx} % for \resizebox
\usepackage{multirow} % for multirow tables
\usepackage{cleveref}
\usepackage{tikz}
\usetikzlibrary{positioning}
\usetikzlibrary{shapes, arrows.meta, positioning}
\tikzstyle{codebox} = [
  rectangle, draw=gray!50, fill=gray!5,
  rounded corners, inner sep=8pt, text width=0.3\linewidth,
  minimum height=2.5cm, align=left
]
\tikzstyle{modelbox} = [
  rectangle, draw=blue!50, fill=blue!10,
  minimum height=2cm,  text width=0.2\linewidth,
  rounded corners, align=center
]
\tikzstyle{arrow} = [thick, ->, >=Stealth]

\DeclareCaptionStyle{ruled}{labelfont=normalfont,labelsep=colon,strut=off} % DO NOT CHANGE THIS
\floatstyle{ruled}
\newfloat{listing}{tb}{lst}{}
\floatname{listing}{Listing}
\nocopyright 
\title{MoSE: Hierarchical Self-Distillation Enhances Early Layer Embeddings}
\author {
    Andrea Gurioli\textsuperscript{\rm 1},
    Federico Pennino\textsuperscript{\rm 1},
    Joao Monteiro\textsuperscript{\rm 2}\thanks{Work done before joining Apple},
   Maurizio Gabbrielli\textsuperscript{\rm 1} 
}
\affiliations {
    \textsuperscript{\rm 1}University of Bologna\\
    \textsuperscript{\rm 2}Apple MLR\\
    andrea.gurioli5@unibo.it, federico.pennino2@unibo.it, joaomonteirof@apple.com,  maurizio.gabbrielli@unibo.it
    
}

\newcommand{\ourmodel}{{\textsc{MoSE}}}
\newcommand{\ourdataset}{{\textsc{SynthCoNL}}}
\newcommand{\sizeDataset}{{\textsc{1,071,367}}}

\begin{document}

\maketitle

\begin{abstract}

% Below's version made with help CLaude:

Deploying language models often requires navigating accuracy vs. performance trade-offs to meet latency constraints while preserving utility. Traditional model distillation reduces size but incurs substantial costs through training separate models. We introduce \textsc{ModularStarEncoder} (\ourmodel{}), a 1-billion-parameter multi-exit encoder for code retrieval and classification that employs a novel Self-Distillation mechanism. This approach significantly enhances lower-layer representations, enabling flexible deployment of different model portions with favorable performance trade-offs. Our architecture improves text-to-code and code-to-code search by targeting specific encoder layers as exit heads, where higher layers guide earlier ones during training, thereby improving intermediate representations at minimal additional cost. We further enhance \ourmodel{} with a repository-level contextual loss that maximizes training context window utilization. Additionally, we release a new dataset created through code translation that extends text-to-code benchmarks with cross-language code-to-code pairs. Evaluations demonstrate the effectiveness of Self-Distillation as a principled approach to trading inference cost for accuracy across various code understanding tasks.
\end{abstract}

\begin{links}
     \link{MoSE}{https://huggingface.co/modularStarEncoder}
     \link{SynthCoNL}{https://huggingface.co/datasets/modularStarEncoder/SynthCoNL}
\end{links}

\section{Introduction}
\label{sec:introduction}
Large language models (LLMs) have significantly impacted natural language processing \citep{empiricalLLM}, but their computational demands pose substantial deployment challenges.  
The field has responded with various strategies aimed at improving efficiency: quantization techniques reduce numerical precision \citep{quantization,quatizationAWQ,quatizationExtreme}; knowledge distillation trains smaller ``student'' models to emulate larger ``teacher'' models \citep{distillBERT,TinyBERT}; and pruning methods eliminate less influential parameters \citep{pruning}. Concurrently, model families like \textsc{LLaMA} \citep{llama3}, \textsc{Qwen} \citep{qwen}, \textsc{Mistral} \citep{mistral}, and \textsc{SmolLM} \citep{allal2025smollm2} exemplify a shift toward more efficient architectures at varying parameter scales.

Dynamic inference approaches further optimize efficiency through mechanisms like multi-exit networks. Early-exit architectures such as \textsc{BranchyNet} \citep{branchynet} balance computation and accuracy by allowing predictions at intermediate layers, while Matryoshka representation learning \citep{matryoshka} enables adjustment of computational complexity through embedding dimensionality pruning. These approaches maintain performance while reducing computational demands in resource-constrained environments. Furthermore, insights from the \textsc{SimCLR} framework~\citep{simCLR} revealed that the final layer of a model does not necessarily produce the most useful representations. \citet{mid_layer_representation} analyzed the geometric evolution of hidden representations within large transformer models across layers, outlining that intermediate layers are identified as holding the most semantically rich representations for downstream tasks.
%%%%%%%%%%%%%%%%%%%%%%%%%%%%%%%%%%%%%
\begin{figure*}[t!]
    \centering
    % Panel (a)
    \begin{subfigure}[t]{0.40\textwidth}
        \centering
        \raisebox{0.3cm}{
        \resizebox{0.9\textwidth}{!}{%
\begin{tikzpicture}[
  % consistent arrow heads
  >={Latex[length=2mm]},
  % styles
  % ========== CHANGE RECTANGLE SIZES HERE ==========
  layer/.style={draw, fill=blue!15, 
                minimum width=3.5cm,      % ← CHANGE WIDTH HERE (layer boxes)
                minimum height=1.1cm,     % ← CHANGE HEIGHT HERE (layer boxes) **INCREASED**
                inner sep=3pt,
                rounded corners=2pt, 
                inner sep=1pt,
                font=\huge\bfseries},    % ← CHANGE FONT SIZE HERE (layer boxes)
  loss/.style={draw, dashed, fill=red!10, align=center, 
               minimum width=3.8cm,       % ← CHANGE WIDTH HERE (loss boxes)
               minimum height=1.8cm,      % ← CHANGE HEIGHT HERE (loss boxes) **INCREASED**
               inner sep=1pt,
               rounded corners=2pt, 
               font=\huge},              % ← CHANGE FONT SIZE HERE (loss boxes)
  sumnode/.style={draw, rectangle, rounded corners=2pt, fill=yellow!10, 
                  minimum size=18mm,      % ← CHANGE SIZE HERE (sum box)
                  inner sep=1pt, 
                  font=\huge},           % ← CHANGE FONT SIZE HERE (sum box)
  ellipsis/.style={font=\huge},          % ← CHANGE FONT SIZE HERE (dots)
  % ========== CHANGE SPACING HERE ==========
  node distance=1mm and 5mm               % ← CHANGE SPACING: vertical and horizontal
]

% Input
\node[layer] (input) {Input};

% Sequential layer blocks
\node[ellipsis, below=of input] (g1) {$\vdots$};
\node[layer, below=of g1] (l4) {Layer 4};
\node[ellipsis, below=of l4] (dots1) {$\vdots$};
\node[layer, below=of dots1] (l9) {Layer 9};
\node[ellipsis, below=of l9] (dots2) {$\vdots$};
\node[layer, below=of dots2] (l18) {Layer 18};
\node[ellipsis, below=of l18] (dots3) {$\vdots$};
\node[layer, below=of dots3] (l27) {Layer 27};
\node[ellipsis, below=of l27] (dots4) {$\vdots$};
\node[layer, below=of dots4] (l36) {Layer 36};

% Loss heads
\def\lossxshift{4mm}                      % ← CHANGE HORIZONTAL SPACE (layers to losses)
\foreach \L/\name/\alphaval in {l4/4/0.11,l9/9/0.25,l18/18/0.50,l27/27/0.75,l36/36/1.0}{%
  \node[loss, right=\lossxshift of \L] (loss\name) {$\mathcal{L}_{\name}$\\ $\alpha=\alphaval$};
  \draw[->] (\L.east) -- (loss\name.west);
}

% Sum node and total loss
\node[sumnode, right=10mm of loss18] (sum) {$\displaystyle\sum \alpha_i \mathcal{L}_i$};
                                          % ↑ CHANGE HORIZONTAL SPACE (losses to sum)

% Connect losses to sum
\foreach \name in {4,9,18,27,36}{%
    \draw[-] (loss\name.east) -- (sum.west);
}

% Vertical pipeline arrows
\foreach \src/\dst in {input/g1,l4/dots1,l9/dots2,l18/dots3,l27/dots4}{%
    \draw[-] (\src.south) -- (\dst.north);
}
\foreach \src/\dst in {g1/l4,dots1/l9,dots2/l18,dots3/l27,dots4/l36}{%
    \draw[->] (\src.south) -- (\dst.north);
}
\end{tikzpicture}
}
        }
        \caption{}
        \label{fig:sub-pipeline}
    \end{subfigure}
    \hfill
    % Panel (b)
    \begin{subfigure}[t]{0.59\textwidth}
        \centering
        \includegraphics[width=\linewidth]{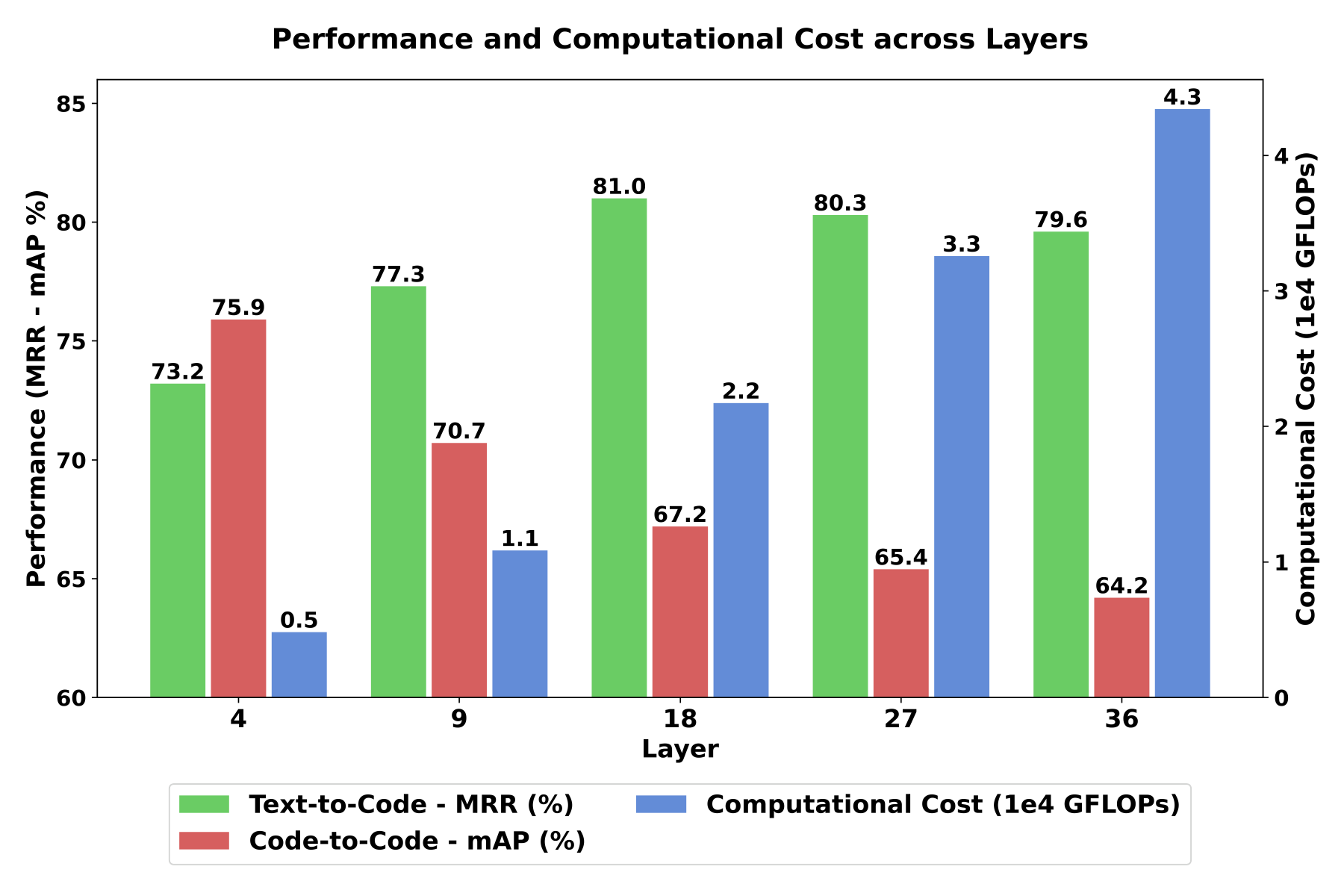}
        \caption{}
        \label{fig:sub-throughput-mrr}
    \end{subfigure}

    \caption{
      \textbf{(a)} Overview of our multi‐exit Self-Distillation encoder, shown here with exit heads at selected layers (e.g., Layers 4, 9, 18, 27, and 36). Each exit head predicts an output embedding and adds a layer loss, contribution weighted by a coefficient $\alpha_i$, summed into the overall objective $\mathcal{L}$.
      \textbf{(b)} Computational cost (GFLOPs) vs.\ performances trade-off at different exit layers over \textsc{CodeSearchNet} dataset (text-to-code in avg. MRR) and POJ104 (code-to-code in mAP). 
      Despite a reduction of approximately 90\% floating point operations from layer 36 to layer 4, MRR performance only drops by 6.4\% in absolute terms. For POJ104, our best results are observed in the initial layers.
    }
    \label{fig:combined}
\end{figure*}
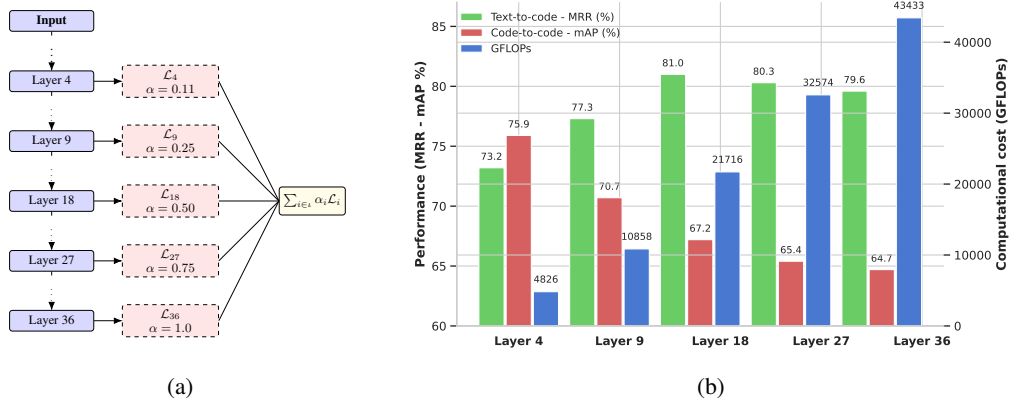

Building on these principles, we introduce \textsc{ModularStarEncoder} (\ourmodel{}), a modular multi-exit encoder built on top of the \textsc{StarCoder-2}~\citep{Starcoder2} architecture, with 1 billion parameters that incorporate intra-model Self-Distillation. As illustrated in Figure~\ref{fig:sub-pipeline}, our architecture applies training objectives at multiple exit points throughout the network (layers 4, 9, 18, 27, and 36), with each exit contributing a weighted loss component to the overall objective. This approach encourages lower layers to learn better representations by mimicking higher-layer outputs.
Figure~\ref{fig:sub-throughput-mrr} demonstrates the resulting accuracy-computation trade-off of encoders trained under this framework: despite a 90\% reduction in floating point operations from layer 36 to layer 4,  performance over text-to-code retrieval drops by only 6.4\% in absolute terms, while the model maintains the best performances for code-to-code retrieval in very early layers.  

Notably, the exit point with the best performance isn't necessarily the final layer, as noted in previous findings due to \citet{simCLR,mid_layer_representation}. This offers both deployment flexibility and a natural model selection mechanism for specific tasks. Users can select which portion of the model to use during inference, starting from the first exit point at 160M parameters. 
By allowing users to select an earlier exit point, \ourmodel{} provides a direct mechanism to decrease computational demands. This flexibility is crucial as smaller models generally require less energy and memory for inference, leading to more sustainable deployment options~\citep{inference_environment}.

We fine-tuned \ourmodel{} on \ourdataset{} (cf. Sec.~\ref{sec:dataset} for details) for text-to-code and code-to-code retrieval, and on \textsc{BigCloneBench} \citep{bigclonebench} for code clone detection, achieving competitive results among open models while maintaining deployment modularity.

Our contributions are as follows:

\begin{enumerate}
  \item A Self-Distillation methodology that trains multiple models within a unified layer stack, reducing redundancy and improving scalability, an approach that could significantly impact LLM training pipelines dependent on multiple model distillations.
  
  \item \ourmodel{},  pre-trained and fine-tuned, with up to 1 billion parameters and five exit points, allowing users to select model size based on memory and computational constraints.
  
  \item \ourdataset{}, a new dataset of \sizeDataset{} natural language-code-code triplets constructed via code translation, expanding text-to-code benchmarks with cross-language code-to-code pairs.
\end{enumerate}

\begin{figure*}
    \centering
    \includegraphics[width=\linewidth]{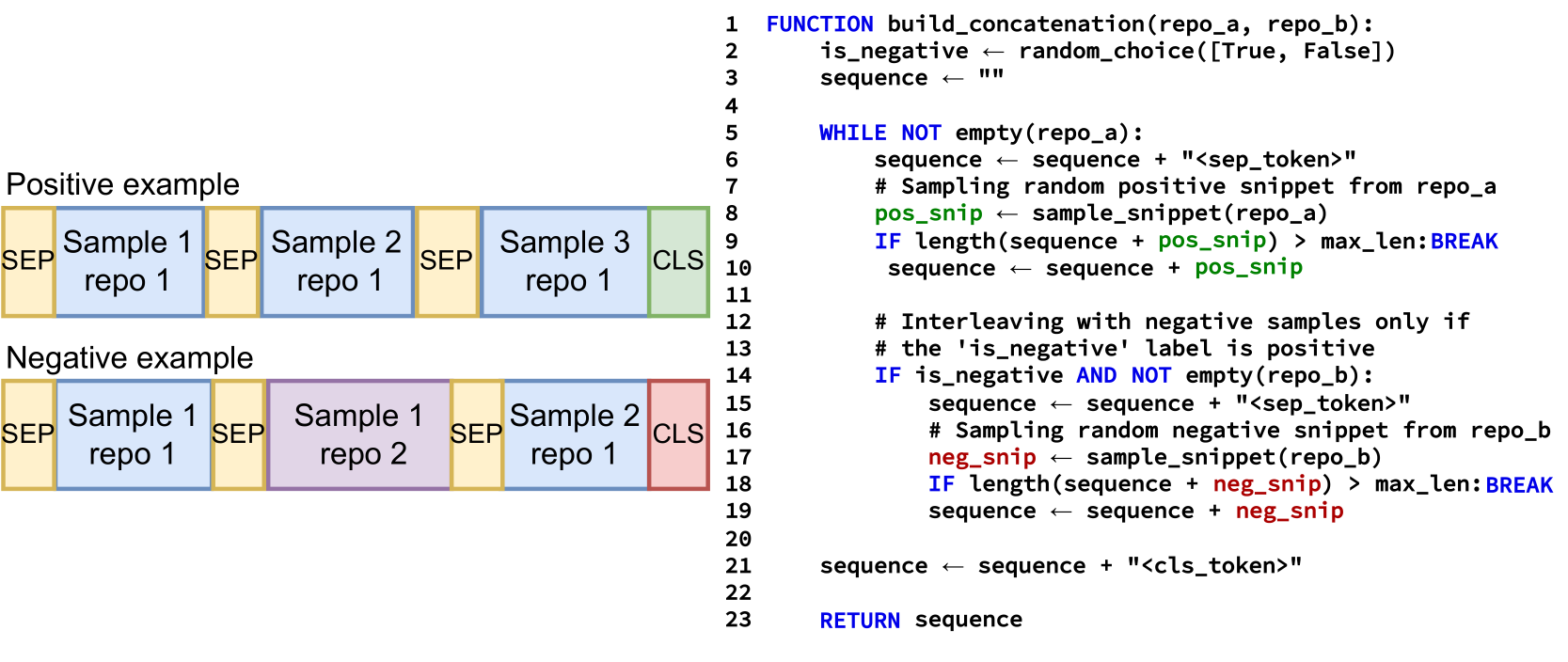}
    \caption{The illustration on the left depicts the in-context loss framework, where samples from various repositories are concatenated. Positive examples originate from the same repository context, whereas negative examples are sourced from different repositories. To enable the model's use of \textsc{FlashAttention V2}, we applied left padding and positioned the CLS token at the end of the sentence. On the right side, you'll find the pseudocode for the in-context loss framework.}
    \label{fig:enter-label}
\end{figure*}

\section{Self-Distillation and In-Context Classification}

As different layers of transformer models capture varying degrees of semantic richness, often with intermediate layers proving most effective for specific tasks~\citep{different_representation_task}, we design our pre-training strategy to support modularity and ease downstream task adaptability. Rather than treating the deepest layer as the sole source of meaningful representations, we aim to enhance representations throughout the network. 

We hypothesize that \emph{applying supervision at multiple depths}, not just at the output, encourages more robust and semantically rich intermediate representations. To this end, we combine multiple training objectives, each contributing to more versatile and generalizable embeddings~\citep{codet5p}. While we retain the widely adopted Masked Language Modeling (MLM) loss~\cite{BERT}, we discard the traditional Next Sentence Prediction (NSP) objective. NSP has been shown to offer minimal benefit after fine-tuning~\citep{roberta}, and it imposes a rigid sentence-level structure that leads to inefficient use of context windows, especially so for long-form code inputs. In its place, we introduce an In-Context Classification (ICC) loss (see Section~\ref{subsec:mlmrcl}) that increases input density, reducing padding and fostering better chunk-level comprehension, essential for downstream tasks such as semantic retrieval or classification.

By integrating these two complementary objectives, MLM and ICC, and distributing their supervision across multiple layers, forming the basis of our \textit{Self-Distillation} mechanism (see Section~\ref{subsec:matr}), our training approach improves internal representation capabilities and flexibility, ultimately yielding a model better suited for modular deployment and early exiting.

% As different layers of transformer models capture varying degrees of semantic richness, we design our pre-training strategy to support modularity and layer-wise adaptability. Rather than treating the deepest layer as the sole source of meaningful representation, we aim to enhance useful representations throughout the network. While we adopt the masked language modeling objective established by \textsc{BERT}~\citep{bertNLP}, our approach is tailored to enhance the model’s internal representations through two novel strategies. First, we introduce an In-Context classification loss aimed at maximizing input density (i.e., minimizing padding), encouraging chunk-level understanding that supports more efficient downstream adaptation (see \cref{subsec:mlmrcl}). Second, to promote modular and robust layer-wise learning, we propose a Multi-Layer loss that instills a form of internal consistency across layers, culminating in our \textit{Self-Distillation} mechanism (detailed in \cref{subsec:matr}). These additions reflect our central goal: to endow early layers with richer semantics, enabling more lightweight and flexible deployment of language models.

\subsection{Masked Language Modeling and In-Context Classification}
\label{subsec:mlmrcl}
% The training objectives of BERT~\citep{BERT}, specifically MLM and Next Sentence Prediction (NSP), have become a de facto standard. However, the NSP loss constrains the context window length to the sentence length, leading to too many padding tokens and redundant computation~\citep{paddedbert}, and has been shown not to yield significant benefits after fine-tuning~\citep{modernbert,aroca2020losses}. Given that the average number of tokens per data sample in \textsc{TheStackV2} is 630, a large context window of 2,048 results in substantial padding, making long-context training inefficient. While \citet{codet5p} demonstrated the advantages of training LLMs with multiple objectives,
We revisited the NSP loss and introduced an In-Context Classification (ICC) objective. We hypothesize that predicting whether multiple code snippets belong to the same context (in our case, the same repository) can enhance semantic search performance while allowing efficient concatenation of multiple code fragments. Our final training objective is the summation of two losses: (1) MLM loss and (2) ICC loss: 
$
    \mathcal{L}=\mathcal{L}_{MLM}+\mathcal{L}_{ICC}
$.

In $\mathcal{L}_{MLM}$, a certain percentage of tokens are randomly masked and predicted using a classification head. Following \citet{sage}, we adopt a 15\% masking rate with the standard 80-10-10 token replacement strategy~\citep{bertNLP}. The secondary objective, $\mathcal{L}_{ICC}$, classifies whether randomly concatenated inputs (separated by a $[\text{SEP}]$ special token) originate from the same repository (see Figure~\ref{fig:enter-label}). Each concatenated sample has a 50\% probability of containing source code from different repositories. This approach increases input density, reducing padding by expanding the average input length from 630 to 1,300 tokens. Since repositories are inherently modular and often contain files written in multiple languages, learning from repository-level context may also improve inter-language understanding.

\begin{table*}[t]
    \resizebox{\textwidth}{!}{\centering
    %\begin{tabular}{l|cccccccc}
    \begin{tabular}{ccccccccc|c|c}

    \toprule
    &   \multicolumn{8}{c}{\bfseries CodeSearchNet} & \multicolumn{1}{c}{\bfseries CT} & \bfseries{POJ104}\\

        \midrule
        Model & Ruby & JS & Go & Python & Java & PHP & avg. MRR & avg. NDCG  & MRR & mAP \\
        \midrule
        \ourmodel{} & 74.1 & \textbf{74.0} & 82.5 & \textbf{92.5} & \textbf{78.7} & \textbf{84.5} & \textbf{81.0} & \textbf{84.2} & \textbf{98.9} & \textbf{75.9}  \\
        \textsc{CodeT5+}  & \textbf{78.0} &71.3 &\textbf{92.7} &75.8& 76.2& 70.1 &77.4   &- & 98.4 &24.5  \\

        \textsc{UniXcoder} & 74.0& 68.4& 91.5& 72.0& 72.6& 67.6& 74.4  & -  & 97.6 &41.0\\
        \textsc{ModernBERT-large} & - &  -& - & - &- & - & - & 59.5 & 93.1 & 27.3 \\
        \hline
        \textsc{OpenAI Embedding} &  84.7& 85.3  & 95.9 & 99.8 & 90.1 & 95.6 & 91.9 & 93.3 & 98.8 & 82.9  \\
        \bottomrule
    \end{tabular}}

    \caption{Performance of different models on text-to-code and code-to-code tasks using the \textsc{CodeXGLUE} benchmark suite, specifically in terms of MRR (on CodeSearchNet and CodeTranslation CT dataset), NDCG, and mAP (on POJ104). On CodeSearchNet, \ourmodel{} surpasses all existing open-source models, achieving an improvement of 3.6\% compared to the best-performing open model, \textsc{CodeT5+}. Furthermore, \ourmodel{} significantly reduces the performance gap between open-source models and closed-source alternatives, such as \textsc{OpenAI text-embedding-3-large} referred to as \textsc{OpenAI Embedding}, and performs comparably on the CT dataset. \ourmodel{} demonstrates competitive results with the CT benchmark and also outperforms all open-source models on the POJ104 benchmark. }

    \label{table:resultst2c}
\end{table*}

\subsection{Self-Distillation: Multi-Layer loss}\label{subsec:matr}

To achieve layer-wise modularity in transformer architectures, we apply the previously introduced loss (Section~\ref{subsec:mlmrcl}) across a selected set of layers, sharing classification heads (masked language modeling and in-context classification) \emph{while incorporating a positional embedding of the layer index}. The total loss is computed as the sum of individual layer losses, weighted by a factor $\alpha$ to prioritize deeper layers:
$
\mathcal{L} = \sum_{i \in \iota}\mathcal{L}_{i}  \cdot \alpha\
$ where $\alpha= i/|I|$ and $I = \{1, \dots, 36\}$ 
represents all layers, and the selected subset  $\iota = \{4, 9, 18, 27, 36\}$ defines the layers where the loss is applied.  The selected subset was chosen to enable four model variants equally spaced in depth (9, 18, 27, 36). We also train a ``tiny'' single-exit baseline to enable isolating the Self-Distillation effect. The \textit{Self-Distillation} mechanism illustrated in \figurename~\ref{fig:combined} allows for flexible model deployment, enabling adaptive layer pruning with minimal impact on performance. 

\section{Pre-training}
\label{sub:arch}
% aims - to reach 1B parameter to reach 2048 of context length, focus on code, an architecture splittable in several sub modules 

We built \ourmodel{} on top of \textsc{StarCoder-2}~\citep{Starcoder2}, applying several modifications to the model. We reduced its original size, resulting in a 1B-parameter model.
Our architecture comprises 36 hidden layers and adopts Grouped Query Attention (GQA)~\citep{gqa} with 16 attention heads and 4 key-value heads. \ourmodel{} relies upon Rotary Positional Encoding (RoPE)~\citep{ROPE} with a base period $\theta = 10^{6}$ and features a hidden dimensionality of 1,024 with an intermediate size of 12,288. 

To make our model class bidirectional, we removed the causal masking from the self-attention operations within \textsc{StarCoder-2}. Aiming for modularity, we also replaced sliding window attention with full attention. This step was taken to avoid the receptive field phenomenon of the sliding window mechanisms~\citep{sliding_window}.
Finally, our implementation integrates \textsc{FlashAttention V2}~\citep{flashAttention2} for faster inference. We have included and summarized additional details about the model architecture in the Appendix.

\subsection{Training Details}
% TODO uncomment for camera READY this version, delete the other one
% We pre-trained \ourmodel{} with a batch size of 3.99M tokens for 245 000 training steps, processing $\approx$1T tokens. We conducted pre-training and fine-tuning on 512 NVIDIA Ampere (64GB) GPUs using the Leonardo supercomputer~\citep{cineca}, requiring 450 000 GPU working hours. 

% We pre-trained \ourmodel{} with a batch size of 3.99M tokens for 245 000 training steps, processing $\approx$1T tokens. We conducted pre-training and fine-tuning on 512 NVIDIA Ampere (64GB) GPUs, requiring 450 000 GPU working hours. 

% To enable both token-level and snippet-level embeddings after pre-training\todo[inline]{??}, we employed a multi-objective pre-training strategy that combined two losses, detailed in \cref{subsec:mlmrcl} and \cref{subsec:matr}. The pre-training was performed on \textsc{TheStackV2}, whose context length analysis revealed an average of $\approx630$ tokens per code snippet. As described in \cref{subsec:mlmrcl}, we concatenated multiple snippets to facilitate our multi-loss methodology, allowing our in-context classification loss to expand the average context window to $\approx1 300$ tokens, reaching the maximum context length 20\% of the time.

We pre-trained \ourmodel{} with a batch size of 4M tokens with a maximum context length of 2,048 tokens, for 245,000 training steps, processing approximately 1T tokens from \textsc{TheStackV2}~\citep{Starcoder2} dataset.

% The pre-training was performed with \textsc{TheStackV2} dataset, whose context length analysis revealed an average of $\approx630$ tokens per code snippet, far below our max context length of 2 048 tokens per sentence. To obtain a dense input, reducing padding tokens, we concatenated multiple snippets to allow our in-context classification loss to expand the average context window to $\approx1 300$ tokens, reaching the maximum context length 20\% of the time (details in \cref{subsec:mlmrcl}). 

We used the AdamW optimizer with $\beta_1$ set to 0.9, $\beta_2$ to 0.95, $\epsilon$ to $1\mathrm{e}{-6}$, and a weight decay of $1\mathrm{e}{-1}$. We initialized the learning rate at $6.24\mathrm{e}{-4}$ and decreased it using a multi-step learning rate scheduler~\citep{deepseek1} with 4,000 warmup steps. The learning rate was reduced at 120,000, 185,000, 220,000, 230,000, and 240,000 training steps, applying a decay factor of 0.36, and from step 185,000 onward, further reduced by factors of 0.1, 0.031, 0.01, and 0.001. 
We conducted pre-training and fine-tuning on 512 NVIDIA Ampere (64GB) GPUs, requiring 450,000 GPU hours.
%on the \textsc{Leonardo} supercomputer~\citep{cineca}. 

\section{Downstream Tasks}
\label{sec:downstream}
\subsection{Fine-tuning for Retrieval, text-to-code and code-to-code search}
\label{retrieval}

Using the \ourdataset{} dataset detailed in Section~\ref{sec:dataset}, we leveraged an instruction prompting approach~\cite{instructioner} and trained a single model for both text-to-code and code-to-code retrieval tasks. The optimization objective uses \textsc{CLIP}-style loss~\citep{clip} with the same multi-layer approach discussed in Section~\ref{subsec:matr}, modified to enhance representation learning.
Accordingly, we replaced the single-head projection of the multi-layer loss with five distinct projection heads, applied at different exit points of the pre-trained model (layers 4, 9, 18, 27, and 36). We used a batch of 2,048 examples, ensuring that text-to-code and code-to-code were equally distributed across the batch.
 
% To enhance representation learning, we replace the standard single-head projection with five distinct projection heads, applied at different exit points of the pre-trained model (layers 4, 9, 18, 27, and 36). We used a batch of 2 048 elements, ensuring that text-to-code and code-to-code were equally distributed across the batch. 

We performed data augmentation by randomly replacing frequently occurring words (appearing more than twice and having at least three characters) with random strings. We applied the augmentation exclusively to code snippets in 30\% of cases, leaving natural language descriptions unchanged. After conducting a grid search, we selected $1\mathrm{e}{-5}$ as the learning rate, maintained throughout the fine-tuning process, and set the temperature parameter at 10.0.

\begin{table}[t]
    \centering
  
    %\resizebox{0.7\textwidth}{!}{\begin{tabular}{l|ccc}
    \resizebox{0.4\textwidth}{!}{\begin{tabular}{cccc}

        \toprule
       &  \multicolumn{3}{c}{\bfseries BigCloneBench} \\

        \midrule
        Model  & Recall & Precision & F1 \\
        \midrule
        \ourmodel{} (\emph{L4})& 96.4 &89.8 & 93.0 \\
        \ourmodel{} (\emph{L9})& \textbf{96.6} &90.4 & 93.4 \\
        \ourmodel{} (\emph{L18})& 96.4 &\textbf{92.1} & \textbf{94.2} \\
        \ourmodel{} (\emph{L27})& 96.5 &91.8 & 94.1 \\
        \ourmodel{} (\emph{L36})& 96.5 &91.7 & 94.1 \\
        \hline
        \textsc{UniXcoder}  & 92.9 & \textbf{97.6} & \textbf{95.2}\\
        \textsc{CodeBERT}  & 94.7& 93.4& 94.1 \\
        \textsc{CodeT5+} (770M)  & \textbf{96.7} &93.5 &95.1 \\

        \bottomrule
    \end{tabular}}

    \caption{\textsc{BigCloneBench} code clone detection performance in a classification setup. \ourmodel{} maintains good performance (\%) throughout different layers (L) with results on par with open models. }
    \label{table:resultsclones}
\end{table}

\subsection{Fine-tuning for classification, code clone detection}
In addition to fine-tuning the model for multiple retrieval tasks as discussed in Section~\ref{retrieval}, we also explored how our methodology performs in a classification setup. We tested how the Multi-Layer and the In-Context classification losses adapt to the code clone detection task by: (1) fine-tuning multiple classification heads, one for each of the five exit points (layers 4, 9, 18, 27, and 36), and (2) leveraging the model’s ability to understand relationships between code snippets by inputting pairs of (hypothetical clone) code segments and training it to classify whether they are clones. We followed the input pattern devised in Section~\ref{subsec:mlmrcl}, formatting the input as follows: $[\text{SEP}]$ \emph{snippet-1} $[\text{SEP}]$ \emph{snippet-2} $[\text{CLS}]$, and then used the final $[\text{CLS}]$ token representation as input to a classifier. We fine-tuned the model with a constant learning rate of $1\mathrm{e}{-5}$ with 2,000 warmup steps, and used a batch size of 64 elements and trained for 14,000 steps.

\section{\ourdataset{}}
\label{sec:dataset}

\begin{figure*}
  \centering
 \begin{tikzpicture}[node distance=1.0cm and 0.5cm]

% Prompt box
\node[codebox] (prompt) {
  \textbf{Prompt} \\
  Translate this ```\texttt{print("Hello World")}''' \\
  from \textcolor{red!70!black}{Python} to \textcolor{green!50!black}{Rust}. \\
  Here is the translated code: 
  
};

% Model box
\node[modelbox, right=of prompt] (model) {
  \textbf{Qwen2.5-Coder-7B-Instruct}
};

% Output box
\node[codebox, right=of model] (output) {
  \textbf{Output} \\
  \texttt{fn main() \{} \\
  \hspace{1em}\texttt{println!("Hello World!");} \\
  \texttt{\}}
};

% Arrows
\draw[arrow] (prompt) -- (model);
\draw[arrow] (model) -- (output);

\end{tikzpicture}
  \caption{ Prompt provided to \textsc{Qwen2.5-Coder-7B-Instruct} for translating a given code snippet ( \texttt{print("Hello World")} in the example) from a source programming language (\texttt{Python}) to a target one (\texttt{Rust}).
    }
  \label{fig:prompt}
\end{figure*}
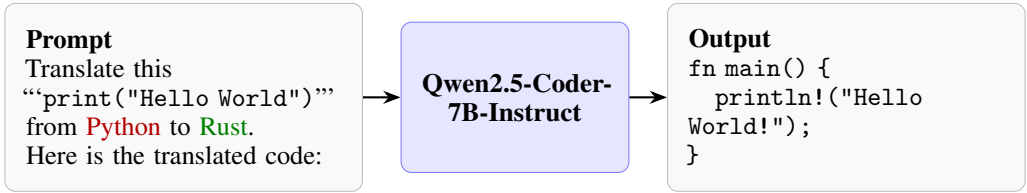

% In the pre-training phase, we leveraged The Stack V2~\cite{Starcoder2}, a large open-source code dataset structured by repository. 

We built \ourdataset{}, a dataset that supports training and evaluation of encoders in text-to-code and code-to-code search, under multiple languages. Using the popular \textsc{CodeSearchNet}~\citep{codesearchnet} as a seed dataset and selecting popular programming languages (Python, Java, Go, and PHP), we augmented it by transpiling available code snippets onto other languages.

To generate semantically similar code snippets for code-to-code search, we translated each snippet into a different language randomly sampled from {Go, Ruby, Python, Java, C++, PHP, C, JavaScript}. We prompted the \textsc{Qwen2.5-Coder-7B-Instruct} model with the source code, the name of the source language, and the name of the target language (see Figure~\ref{fig:prompt}). During code translation, we carried out a greedy sampling so as to prevent semantic discrepancies. 
This process yielded pairs of code snippets in distinct languages tied to the same natural language description. As a result, every sample in the fine-tuning dataset includes a natural language description and two code snippets from distinct languages.  \ourdataset{} contains \sizeDataset{} triplets where, in the first code column, we directly sampled code snippets from \textsc{CodeSearchNet}, including Python, Java, PHP, and Go. The second code column, artificially synthesized via code translation, includes Go, Ruby, JavaScript, Python, C++, PHP, C, and Java code snippets. 
After a manual inspection, we discovered that both columns contained code snippets that differed only in identifiers or function arguments. Several tasks were semantically identical but paraphrased with different parameter requirements (e.g., two identical paraphrased tasks asked for opening a socket on a different port).
During the preprocessing phase of \ourdataset{}, motivated by the dataset's redundancy and preliminary experiments that show its effectiveness on the model's performance, we near-deduplicated the dataset using both the \textsc{CodeSearchNet} code column and the synthesized code column. During the data near deduplication phase, we relied on Locality Sensitive Hashing (LSH) with a Jaccard similarity threshold of 0.7 and 256 permutations, analyzing character-level 5-grams.

\section{Results and Discussion}

\subsection{Evaluation}
\label{sec:benchmarks}

We evaluated \ourmodel{}, after fine-tuned for retrieval, on diverse code understanding tasks, leveraging the \textsc{CodeXGLUE}~\citep{codexglue} group of benchmarks. 

On \textsc{CodeSearchNet} dataset, we assessed cross-modal (natural language to code) learning by retrieving target code from 999 distractors using natural language queries. For code-to-code retrieval, again against 999 distractors, we utilized \textsc{CodeXGLUE}'s Code Translation (CT) dataset to test cross-lingual retrieval (e.g., Java to C\# snippets implementing the same functionality), and its \textsc{POJ104} dataset (C++ snippets, semantically equivalent but syntactically different) for intra-lingual semantic search, evaluating generalization over structural differences while preserving semantics. In our retrieval performance evaluation, when comparing \ourmodel{} with other encoders, we included results from \textsc{OpenAI’s text-embedding-3-large} as a representative commercial, closed-source model. 

Additionally, \ourmodel{} was fine-tuned and evaluated on the separate \textsc{BigCloneBench}~\citep{bigclonebench}
dataset for Java code clone detection (a binary classification task), assessing its classification performance.

\subsection{Benchmarks}
Table~\ref{table:resultst2c} presents the results for the \textsc{CodeSearchNet} (text-to-code) task in terms of Mean Reciprocal Rank (MRR, higher is better) for each single language, average Normalized Discounted Cumulative Gain (NDCG, higher is better), and average MRR.
Results for \textsc{UniXcoder}, \textsc{ModernBERT}, and \textsc{CodeT5+} are reported from the original papers~\citep{unixcoder,modernbert,codet5p}. On \textsc{CodeSearchNet}, \ourmodel{} achieves an MRR of 81.0 and a NDCG of 84.2, outperforming \textsc{CodeT5+} (770M), \textsc{UniXcoder}, and \textsc{ModernBERT‐large}. The only encoder that surpasses \ourmodel{} is \textsc{OpenAI’s text-embedding-3-large}. 

In Table~\ref{table:resultst2c}, we also present results from both POJ104 and CT datasets (code-to-code) reported respectively in terms of MRR for code translation (Java to C\# retrieval) and mean average precision for POJ104 (C++ to C++ retrieval). \ourmodel{} reaches the best performance among the evaluated models. We further replicated the benchmarking for all models in a zero-shot setting for code-to-code tasks, as our model does not integrate POJ104 and the code translation datasets in the training set.
We present a comprehensive and detailed analysis of code retrieval errors in the Appendix.

\begin{figure}[t!]
  \centering
  \includegraphics[width=0.43\textwidth]{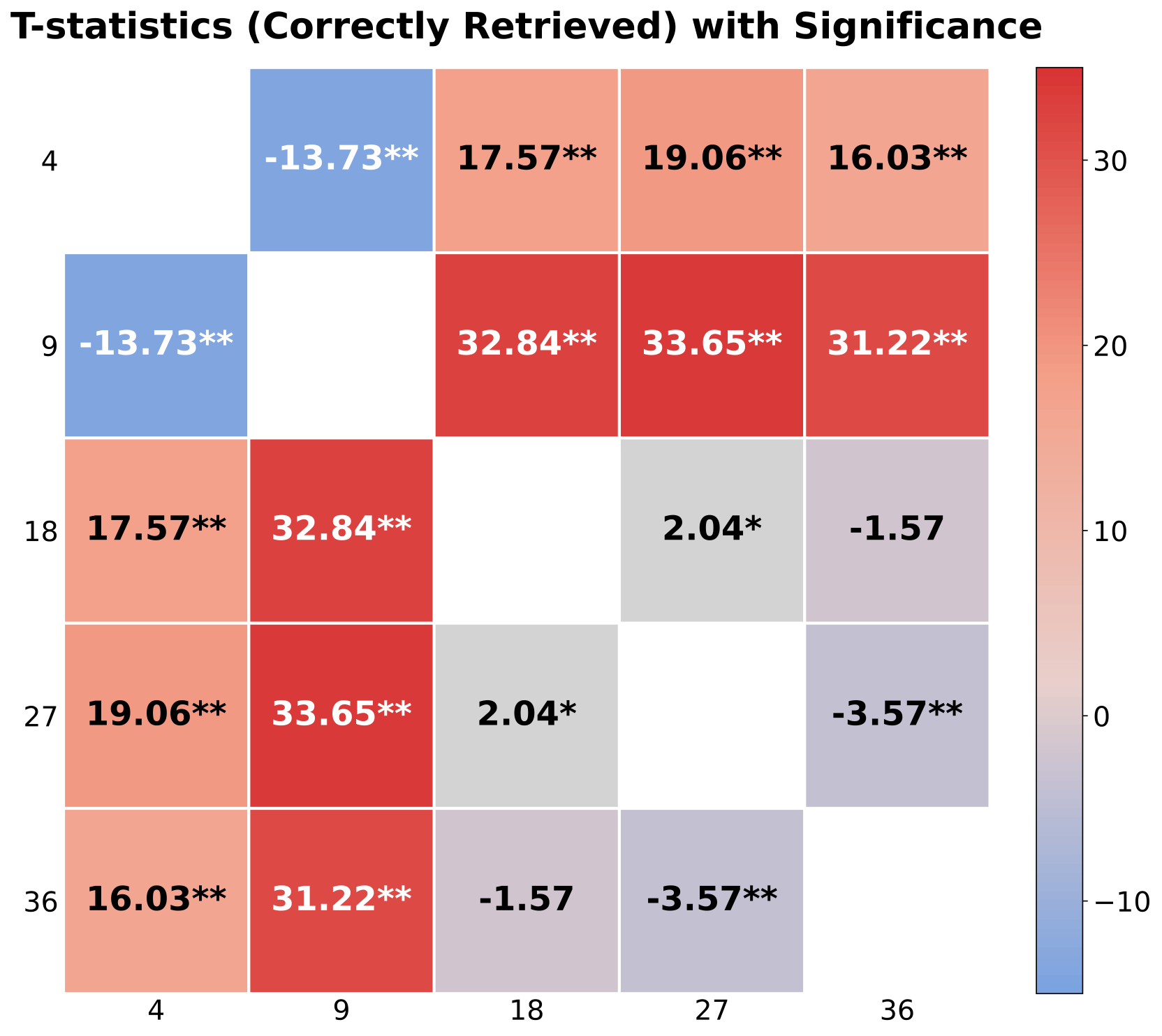}
  \caption{Similarity score Heatmap presenting results from permutation tests (10,000 permutations with $\alpha$ = 0.05) across different exit points for  Python-to-Java retrieval.  * indicates p $<$ 0.05 and ** indicates p $<$ 0.001. Different layers, despite a common training objective, yield different similarity scores.}
  \label{fig:sideplot}
\end{figure}
On the \textsc{POJ104} dataset, in zero-shot, \ourmodel{} achieves a mean Average Precision (mAP, higher is better) of 75.9, which is the highest among open-source models to our knowledge. However, it is significantly behind \textsc{OpenAI text-embedding-3-large}. We underscore that a direct comparison with \textsc{OpenAI text‐embedding‐3‐large} remains challenging because it is closed‐source, and details such as model size, training methodology, or potential test data leakage are undisclosed.
We observe a difference in performance profiles depending on whether the task is framed as text-to-code or code-to-code retrieval. As shown in Figure~\ref{fig:sub-throughput-mrr}, shallow layers yield better performance in the code-to-code setting, whereas for text-to-code retrieval, optimal performance emerges at layer 18. 

This representation divergence suggests that the input modality influences where semantic alignment is most effectively captured within the model. This observation aligns with \citet{mid_layer_representation} findings, underlining that \textit{semantic information is distributed unevenly across layers} and is better expressed in the intermediate ones. \ourmodel{} leverages this property by supporting dynamic exit points, enabling users to balance computational cost and task performance based on where the most relevant representations emerge.

To further investigate this phenomenon, we compared similarity scores across layers to assess how each layer represents code snippets differently. Figure~\ref{fig:sideplot} shows permutation test results for a Python-to-Java retrieval task, highlighting significant variations in similarity scores that support our previous findings. A detailed analysis of these results is provided in the Appendix.

In Table~\ref{table:resultsclones}, we report classification performance across different exit layers of \ourmodel{} for the \textsc{BigCloneBench} benchmark. Results show stable performance throughout layers, with F1-scores from Layer 18 onward (94.2–94.1) closely matching those of deeper exits. Earlier layers (e.g., Layers 4 and 9) exhibit slightly lower precision, reflecting the expected trade-off between early inference and representational depth.

Our method remains competitive compared to open models. While \textsc{CodeT5+} (770M) and \textsc{UniXcoder} achieve peaks F1 of 95.1 and 95.2, \ourmodel{} delivers on-par results (-0.9) by Layer 18, with the added benefit of a modular architecture. 

\subsection{Ablation Study}
\label{sec:ablation}
\subsubsection{Effectiveness of the Multi Layer Loss}
\begin{figure*}[t!]%
    \centering
    \subfloat[\centering MRR $\times$ Embedding depth]{\includegraphics[width=0.47\textwidth]{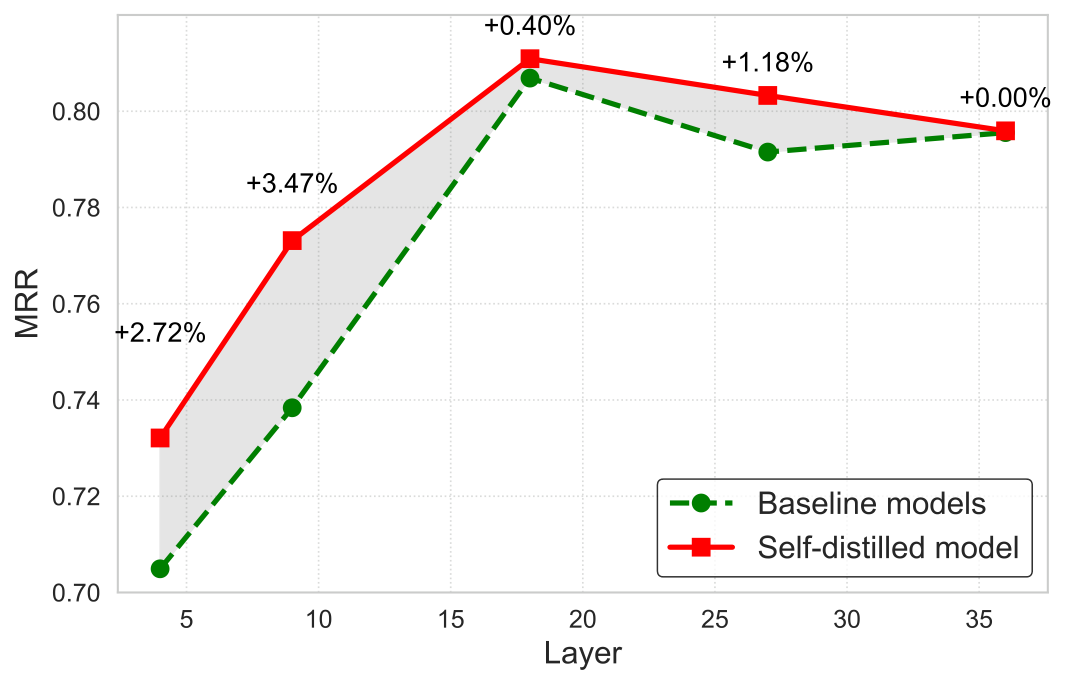}}%
    \qquad
    \subfloat[\centering Recall@1 $\times$ Embedding depth]{{\includegraphics[width=0.47\textwidth]{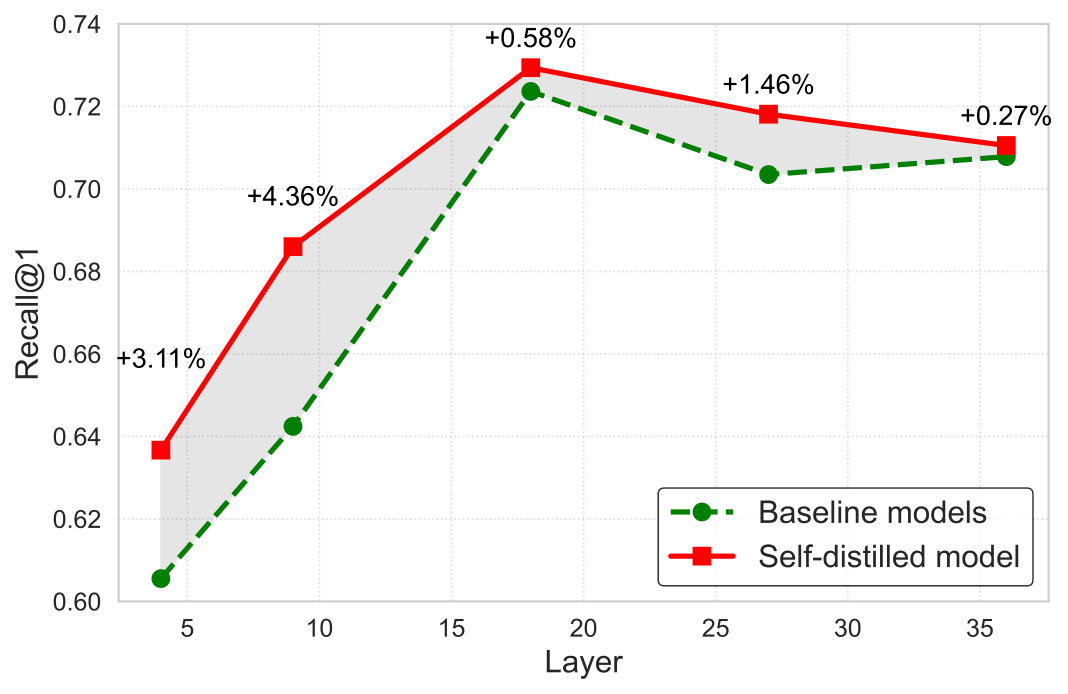} }}%
    \caption{Performance Comparison between \ourmodel{} fine-tuned with multi-layer loss (self-distilled model) and baselines fine-tuned just on single exit points: The graph illustrates averaged MRR and Recall@1 results for different layers over the \textsc{CodeSearchNet} test set. Layers fine-tuned with Multi-Layer loss outperforms the baselines (up to  +4.36\% Recall@1).}%
    \label{fig:self_distillation}%
\end{figure*}

We conducted an ablation study for the multi-layer loss, by fine-tuning each exit point independently on \ourdataset{}, starting from the pre-trained \ourmodel{}. For example, for the baseline on layer 18, we retain the first 18 layers and fine-tune the model using just the projection head on that layer. Finally, we compared the baseline models with the corresponding results (Self-Distilled) of \ourmodel{} fine-tuned for retrieval with the multi-layer loss. All tests are conducted on the \textsc{CodeSearchNet} dataset. \ourmodel{} consistently outperforms the single‐exit baselines (up to +4.36\%  averaged Recall@1 gain on layer 9), indicating that \emph{lower‐level layers benefit from training signals propagated from deeper layers}. This behavior is highlighted in Figure~\ref{fig:self_distillation} where \ourmodel{}, indicated as \emph{Self-Distilled}, outperforms all the single exit baselines consistently. This finding underscores a promising new direction in Self‐Distillation for large‐scale code and text models, enabling high performance even in more compact configurations. Moreover, Figure~\ref{fig:self_distillation} also illustrates that \ourmodel{} maintains robust performance from layers 18 to 36, allowing users to scale down the network to match their memory, computational, or latency constraints while preserving strong retrieval performances. 

\subsubsection{Effectiveness of the In-Context Classification Objective}

\begin{table}[t]
    \centering
  
    %\resizebox{0.7\textwidth}{!}{\begin{tabular}{l|ccc}
    \resizebox{0.45\textwidth}{!}{\begin{tabular}{ccccc}

        \toprule
       &  \multicolumn{3}{c}{\bfseries CoIR (NDCG@10)} \\

        \midrule
        Model  & CSN‑CCR & CoSQA & CT‑DL & CT‑Contest \\
        \midrule
        \ourmodel{} (\emph{ICC})& \textbf{10.1} &\textbf{ 0.4} & \textbf{32.0} &\textbf{12.1} \\
        \ourmodel{} (\emph{NSP})& 6.2 &0.4 & 31.0&12.0 \\

        \bottomrule
    \end{tabular}}
    \caption{Ablation study comparing pre-training objectives (In-Context Classification vs. Next Sentence Prediction), evaluated with NDCG@10 (\%) on the CoIR suite. ICC demonstrates  gains on the cross-context \textsc{CSN‑CCR} task.}

    \label{table:resultsablationContext}
\end{table}
We conducted an ablation study, detailed in Table~\ref{table:resultsablationContext}, to assess whether our In-Context Classification (ICC) loss promotes inter-language understanding. For this study, we pre-trained two small versions of the model (each with 30M parameters) on 3B tokens. We compared the ICC loss with the Next Sentence Prediction (NSP) loss, commonly used in many state-of-the-art encoders~\cite{starcoder,BERT}.

To assess the performance of our models, we utilized the inter-domain section of the CoIR evaluation suite~\cite{CoIR}, which focuses on tasks across various languages and domains. Our results indicated that the In-Context Classification loss either outperformed or matched the effectiveness of the Next Sentence Prediction loss, while providing a more efficient and dense representation.
These results suggest that ICC may serve as a stronger default pretraining signal than NSP for multilingual and cross-context information retrieval tasks.

\section{Related Work}
Since the introduction of ELMo~\citep{elmo}, deep contextual information has enhanced generating embeddings for textual retrieval or classification, reaching state-of-the-art results in several tasks. BERT~\citep{BERT} followed those findings, adapting the Transformer architecture~\citep{attentionIsAllYouNeed} to enable a bi-directional representation with two different training objectives, namely the masked language modeling and the next sentence prediction losses. 
\cite{ALBERT} and \cite{roberta} adapted the BERT architecture to obtain an enhanced pre-trained model by removing or modifying the NSP, focusing on pre-training data or hyperparameters optimization. More recently, \textsc{modernBERT}~\citep{modernbert} tied the gap between modern decoders~\citep{mistral,qwen,llama3, llama, Starcoder2} advancements that rely upon models with an increased number of parameters, trained upon more tokens,  and being capable of handling longer contextual information.

In code representation, large language models must be adapted by training them on a curated corpus focused on software and by leveraging code's syntactic and semantic structures, which differ significantly from natural language.
\citet{codebert} adapted the BERT architecture to produce semantically meaningful embeddings for source code, resulting in \textsc{codeBERT}. This was accomplished by including more source code in the training set and focusing on a training loss that can leverage bimodal (natural language and code) contextual information~\citep{lossCodebert}. \textsc{GraphCodeBERT} enhanced \textsc{codeBERT}~\citep{codebert} representations by incorporating data flow graphs, capturing dependencies between variables and operations, and improving tasks like code summarization and clone detection.
\textsc{UniXcoder}~\citep{unixcoder} extended this by introducing a unified encoder-decoder framework, integrating abstract syntax trees (ASTs) and data flow information. 
\citet{codet5p} expanded these findings with \textsc{CodeT5+}, stressing how multiple losses that leverage code semantics impact the model pertaining. The work incorporated text-code contrastive learning, text-code Matching, and text-code causal LM for better code understanding and generation.
When trying to achieve better performance, research has shifted toward models with a high number of parameters. While this trend appears effective from a performance perspective, end users may face computational or memory limitations as LLMs vary from millions to billions of parameters. 
\citet{distillBERT} pioneered the introduction of knowledge distillation, using a ``teacher'' model that guides a smaller model to emulate its behavior. 
This methodology has been widely adopted and improved upon recently~\citep{deepseekr1,qwen}, becoming a standard for obtaining high-performing smaller LLMs. 

Our work differs from previous work by adapting a modern architecture (\textsc{StarCoder-2}~\citep{Starcoder2}) to a code encoder-only based model and introducing a novel 'Self-Distillation' mechanism. We replace the next sentence prediction loss with an In-Context Classification focused on the repository level and expand the context to 2,048 tokens. Our novel Self-Distillation mechanism improves low-level layers, resulting in a modular transformer architecture without additional teacher models or further data for distillation.

\section{Conclusion}
\label{sec:conclusion}
We introduced \ourmodel{}, a modular, multi-exit encoder that uses Self-Distillation to enhance early-layer representations while achieving competitive results on code understanding benchmarks.
By integrating multiple exit points, \ourmodel{} enables explicit efficiency-accuracy trade-offs tailored to deployment constraints. Our approach not only allows for modular use but also enhances the alignment between task requirements and representation depth. This improves task-specific layer specialization, providing the end-user with a more adaptable model.
Future work will explore combining exit-points and further understanding task type and depth interactions. While our results are encouraging, we acknowledge that computational constraints limited extensive hyperparameter tuning and pre-training ablations, and that fine-tuning with synthetic code generated from code translation has an unclear impact.

\section*{Acknowledgments}
We acknowledge ISCRA for awarding this project access to the LEONARDO supercomputer, owned by the EuroHPC Joint Undertaking, hosted by CINECA (Italy). This work was supported by Future AI Research (FAIR) PE01, SPOKE 8 on PERVASIVE AI funded by the National Recovery and Resilience Plan (NRRP).

\bibliography{aaai2026}


\clearpage

\end{document}